\documentclass[conference]{IEEEtran}
\IEEEoverridecommandlockouts
\makeatletter
\let\NAT@parse\undefined
\makeatother
\usepackage{tcolorbox}
\usepackage{subcaption}
\usepackage{textcomp}
\usepackage{xcolor}
\usepackage{tabularx}
\usepackage{booktabs}
\usepackage{multirow}
\usepackage{csquotes}
\usepackage{booktabs} 
\usepackage{makecell}
\usepackage{threeparttable, tablefootnote}
\usepackage{graphicx}
\usepackage{amsmath,amssymb,amsfonts}
\usepackage{algorithmic}
\usepackage{comment}
\usepackage{balance}
\usepackage{pdflscape,booktabs,tabularx,ragged2e}
\def\BibTeX{{\rm B\kern-.05em{\sc i\kern-.025em b}\kern-.08em
    T\kern-.1667em\lower.7ex\hbox{E}\kern-.125emX}}
\begin{document}
\title{\LARGE \bf
Enhancing Dimension-Reduced Scatter Plots with Class and Feature Centroids}
\author{ Daniel B. Hier, Tayo Obafemi-Ajayi, Gayla R. Olbricht, Devin M. Burns, Sasha Petrenko, Donald C. Wunsch II
        \thanks{Daniel Hier, Gayla Olbricht, Sasha Petrenko, Devin Burns, and Donald Wunsch II are with Missouri University of Science \& Technology (Missouri S \& T, Rolla, MO, USA. Email: \{hierd, olbrichtg, burnsde,  petrenkos, dwunsch\}@mst.edu}
        \thanks{Tayo Obafemi-Ajayi is with Missouri State University, Springfield, MO, USA. Email: tayoobafemiajayi@missouristate.edu}
        \thanks{ Partial support from the Missouri S \& T Kummer Institute, the Mary Finley Endowment, and the Missouri S \& T Intelligent Systems Center are gratefully acknowledged.}
        }
\maketitle
\thispagestyle{empty} 
\pagestyle{empty} 
\begin{abstract}
    Dimension reduction is increasingly applied to high-dimensional biomedical data to improve its interpretability. When datasets are reduced to two dimensions, each observation is assigned an \textit{x} and \textit{y} coordinates and is represented as a point on a scatter plot. A significant challenge lies in interpreting the meaning of the x- and y-axes due to the complexities inherent in dimension reduction. This study addresses this challenge by using the x and y coordinates derived from dimension reduction to calculate class and feature centroids, which can be overlaid onto the scatter plots. This method connects the low-dimension space to the original high-dimensional space. We illustrate the utility of this approach with data derived from the phenotypes of three neurogenetic diseases and demonstrate how the addition of class and feature centroids increases the interpretability of scatter plots.

    \textit{\textbf{Clinical relevance:}}
    Overlaying class and feature centroids onto two-dimensional scatter plots derived from high-dimensional biomedical datasets enhances their interpretability.
    
\end{abstract}
\section{Introduction}
\label{s:introduction}
Big data and precision medicine initiatives are creating large datasets with numerous observations and many features \cite{hulsen2019big,misra2019integrated}. In particular, 
genomic and molecular biological datasets are characterized by high dimensionality \cite{hua2009performance,clarke2008properties}.  
The goal of dimension reduction is to simplify the dataset by reducing the number of features while retaining as much of the original information as possible.  Dimension reduction is especially important when dealing with high-dimensional data, where a large number of features can lead to complexity, increased computational costs, and the risk of overfitting by machine learning models \cite{colgan2013analysis,al2019computational,de2019dimensionality}.  
\begin{figure}
    \centering
    \includegraphics[width=0.49\textwidth]{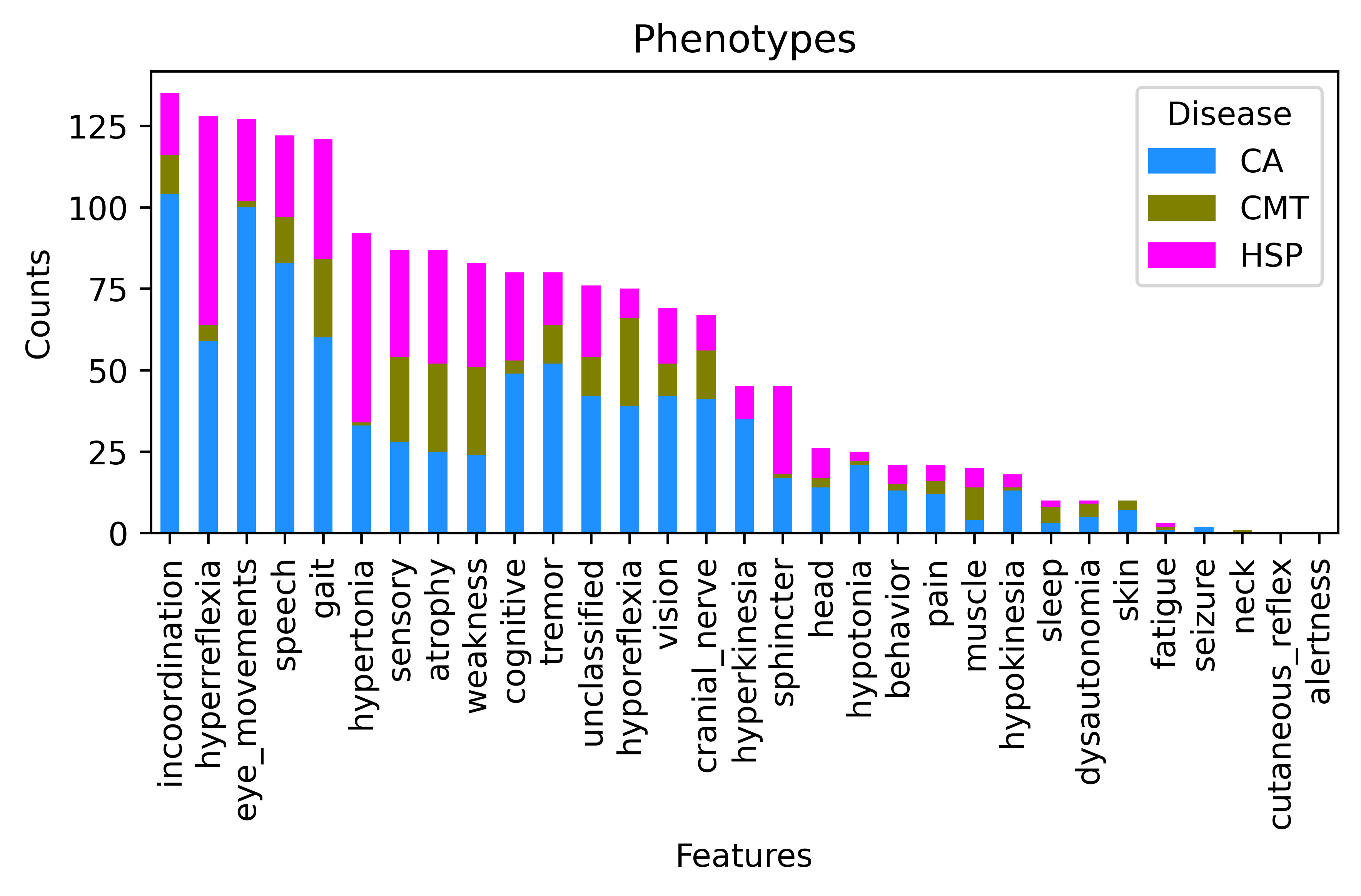}
     \captionsetup{font=footnotesize}
    \caption{
        \footnotesize Stacked bar chart of phenotypes (features) by disease type. Each column shows the frequency of one of the 31 available phenotype superclasses. The three most frequent phenotypes were incoordination, hyperreflexia, and eye movement abnormalities.}
    \label{fig:bar_chart_neurogenetic}
\end{figure}High-dimensional datasets can be visualized as scatter plots when reduced to two dimensions. A variety of dimensionality reduction methods are available, including locally linear embedding (LLE), isometric feature mapping (ISOMAP), multidimensional scaling (MDS), uniform manifold approximation and projection (UMAP), principal component analysis (PCA), and t-distributed stochastic neighbor embedding (t-SNE) \cite{ghojogh2020multidimensional,kobak2019art,linderman2019fast,wattenberg2016use,anowar2021conceptual,ayesha2020overview}. These methods vary in stochasticity, linearity, sensitivity to parameter changes, and their ability to preserve the local and global structure of the data. The interpretation of scatter plots after dimension reduction remains complex. After dimension reduction, the \textit{x} and \textit{y} axes cannot be easily labeled and are typically expressed in arbitrary units that reflect the operations of complex algorithms \cite{rudin2022interpretable}.

We present a method to enhance the interpretability of two-dimensional scatter plots obtained after dimension reduction. Feature and class centroids are overlaid onto the plots to enhance interpretability. This approach addresses the challenge of interpreting the axes of dimension-reduced plots.

To demonstrate the feasibility of this approach, we used it to visualize the phenotypes of three rare neurogenetic diseases: hereditary spastic paraparesis (HSP), hereditary cerebellar ataxia (CA), and Charcot-Marie-Tooth disease (CMT).
Gene mutations cause all \cite{vallat2016classifications,lallemant2021clinical,pareyson2009diagnosis}. We sought to capture the phenotypic variability between these diseases with a scatter plot. Specifically, we reduced a high-dimensional space with 31 available phenotype categories (Fig. \ref{fig:bar_chart_neurogenetic}) to two dimensions and overlaid class and feature centroids. Before dimension reduction, the data had 31 columns (each column representing a phenotype feature) and 235 rows (each row representing a variant of either Charcot-Marie-Tooth disease, cerebellar ataxia, or hereditary spastic paraparesis).

\begin{figure}
\centering
\includegraphics[width=0.45\textwidth]{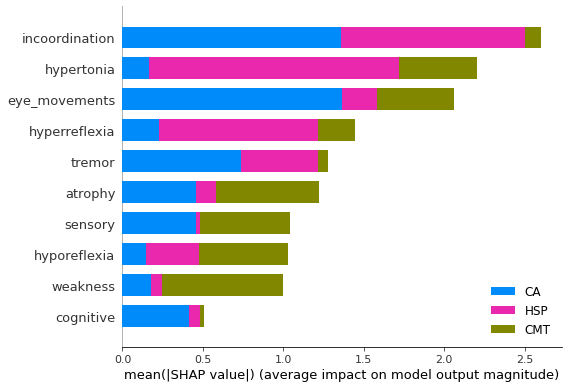}
    \captionsetup{font=footnotesize}
    \caption{
        \footnotesize SHAP values used to identify the most important features predicting class membership.
        Bar length is proportional to influence on a class i.e. incoordination favored cerebellar ataxia; hyporeflexia: Charcot-Marie-Tooth disease; and hypertonia: hereditary spastic paraparesis.
        Top $10$ features were retained for further analysis.}
    \label{fig:SHAP}
\end{figure}
\begin{figure}
    \centering
    \includegraphics[width=0.47\textwidth]{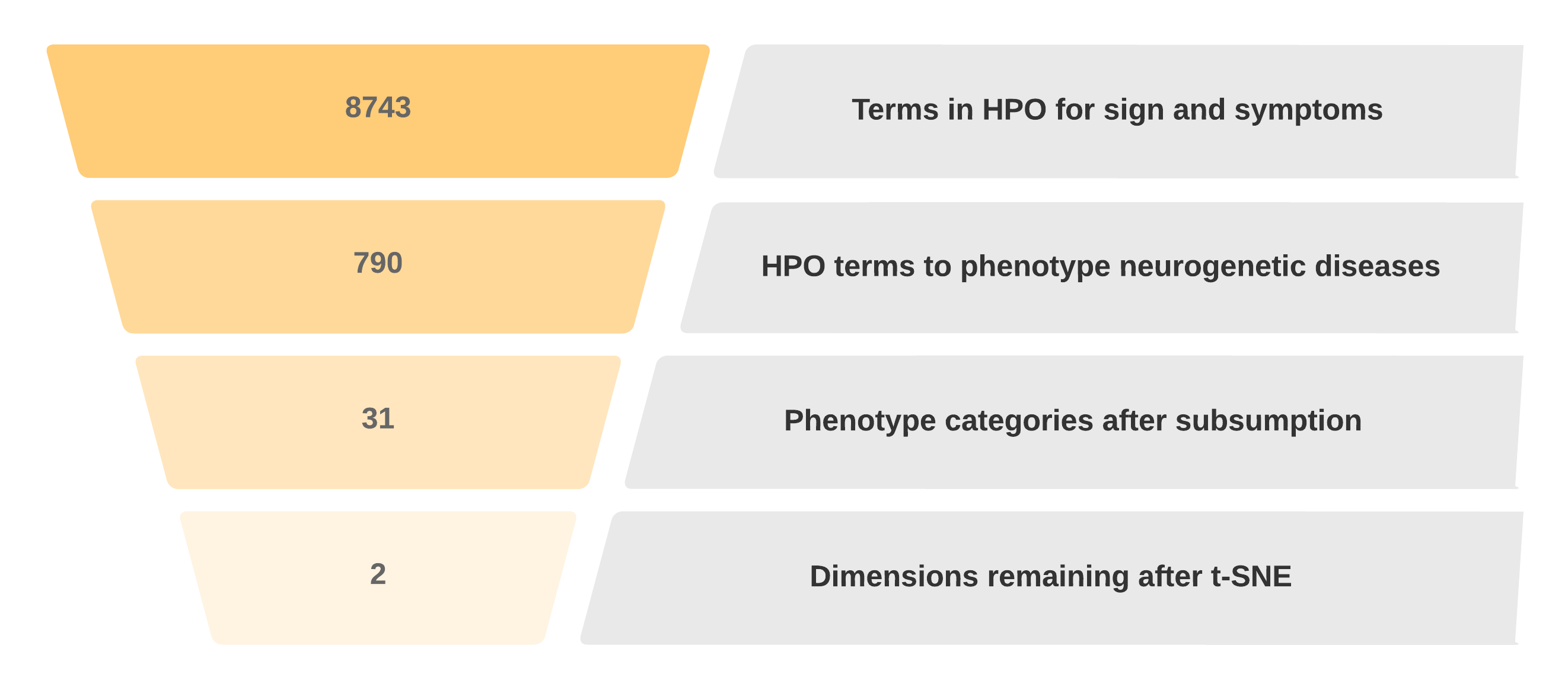}
    \caption{Summary of dimension reduction strategy. The Human Phenotype Ontology (HPO) has 8743 terms; the 235 neurogenetic disease cases used 790 of these terms; subsumption reduced the phenotypes to 31 categories; UMAP and t-SNE reduced the dimensions to 2.}
    \label{fig:Subsumption}
    
\end{figure}\textit{Relevant Prior Work}.
Reviews of dimension reduction methods for the visualization of high-dimensional datasets are available \cite{ wang2021understanding, anowar2021conceptual, ortigossa2022getting, ghojogh2020multidimensional, ayesha2020overview}.
Based on quantitative benchmarks and end-user preferences, \cite{xia2021revisiting} found that t-SNE and UMAP outperformed other dimension reduction methods. \textit{UnProjection} \cite{espadoto2021unprojection} is a project that uses neural networks to reconnect the low-dimension space to the high-dimension space. \textit{Interaxis} is a project that allows end users to interactively manipulate the axes of low-dimensional scatter plots to explore the influence of features on the plotted positions of markers \cite{kim2015interaxis}. In similar work, Eckelt et al. \cite{eckelt2022visual} have suggested adding group centroids to low-dimensional embeddings of high-dimensional datasets. \cite{xu2014visualization} has used annotations to make disease-phenotype plots based on t-SNE more interpretable. \cite{chatzimparmpas2020t} have created \textit{t-viSNE} as a visual interactive inspector of t-SNE scatter plots. It can generate interpretive feature bar charts based on the t-SNE plots that are the circumscribed areas of interest. \cite{sohns2021attribute} introduced \textit{NoLiES}, an interactive system to improve the interpretability of low-dimensional embeddings by circumscribing \textit{rangesets} on the plots.
\section{Methods}
\label{s:methods}
\textit{Data acquisition}.
The variants belonging to the three neurogenetic diseases (Charcot-Marie-Tooth disease, hereditary cerebellar ataxia, and hereditary spastic paraplegia) were extracted from Orphadata \cite{maiella2013orphanet}, resulting in a total of $235$ variants.
The phenotypes for each variants were coded as terms from the Human Phenotype Ontology (HPO) \cite{robinson2012deep}. Nine hundred and seventy unique HPO terms were needed to describe the phenotypes of the 235 variants of neurogenetic diseases (Fig. \ref{fig:Subsumption}. Only HPO terms that were signs or symptoms of neurogenetic disease were retained. Radiological and laboratory findings were excluded. Each HPO term was binary coded (present or absent), creating an array with $235$ rows (each row was a variant of one of the three neurogenetic diseases), two columns of labels (disease name and variant name), and $970$ columns of features (each a phenotype description that was binary coded as 0 or 1 for \textit{absent} or \textit{present}). For example, Row 1 had a disease name of \textit{Charcot-Marie-Tooth disease}, a variant name of \textit{Charcot-Marie-Tooth disease type 1A}, and phenotype features of {hyporeflexia}, {gait disturbance}, {pes cavus}, {distal muscle weakness}, {distal sensory impairment}, {skeletal muscle atrophy}, {sensory ataxia}, gait imbalance, {kyphoscoliosis}, {paresthesia}, {calf muscle hypertrophy}, {diaphragmatic weakness},  {spontaneous pain sensation}, {shoulder pain}, and {hyperactive deep tendon reflexes} (all coded 1 for \textit{present}). Among the $235$ neurogenetic diseases were $125$ variants of cerebellar ataxia (CA), $33$ variants of Charcot-Marie-Tooth disease (CMT), and $77$ variants of hereditary spastic paraparesis (HSP).	

\textit{Dimension reduction.} The number of phenotype features (HPO terms) was reduced from 970 to 31 by the reduction by subsumption method \cite{wunsch2021subsumption, hier2023visualization}. This method is applicable here because of the hierarchical structure of the HPO. Most terminology ontologies (HPO included) are based on a \textit{subsumptive containment hierarchy} with classes hierarchically organized from the general to the specific. Each child class inherits properties from its parent class. This inheritance of properties from a parent is called \textit{subsumption}. 
The reduction by subsumption procedure replaces terms that are highly specific in meaning and lower in the ontology hierarchy with more general terms higher in the hierarchy. For example, a more general term such as  \textit{hyporeflexia} can serve as a container for more specific HPO terms such as areflexia, lower limb hyporeflexia, areflexia of the lower limbs, reduced tendon reflexes, absent Achilles reflex, hyporeflexia of the upper limbs, and hyporeflexia of the lower limbs that are \textit{subsumed} by the category of hyporeflexia.  After replacing more specific phenotype terms with more general phenotype categories by subsumption (Fig. \ref{fig:Subsumption}), the dimension-reduced data array had $235$ rows, two label columns, and $31$ feature columns. The data array was reduced to two dimensions using t-SNE \cite{van2008visualizing} (perplexity = 50, metric = euclidean).

\textit{Calculation of class and feature centroids}.
Dimension reduction by t-SNE created \textit{x}, \textit{y} coordinates for each observation. The \textit{x} and \textit{y} coordinates were concatenated to the original $235 \times 33$ data matrix, yielding a $235 \times 35$ matrix. The concatenated data matrix was sorted and filtered by either the variant name (\textit{class}) column or any of the $31$ phenotype (\textit{feature}) columns. We calculated class centroids (Equation \ref{equation_class}) for the three ground truth classes (cerebellar ataxia, Charcot-Marie-Tooth disease, and hereditary spinal paraparesis).  We calculated feature centroids for the 31 feature categories (Equation \ref{equation_feature}). For plotting purposes, we looked for the most important features that predicted class membership.  We fitted an \textit{xgboost } classifier \cite{Chen_xgboost} to the data and used the Shapley Additive exPlanations (SHAP) values \cite{lundberg2017unified} to find the most important phenotype features for predicting class membership (Fig. \ref{fig:SHAP}). SHAP applies game theory to identify features most predictive of class membership .

\textit{Scatter plots}.
Based on their \textit{x}, \textit{y} coordinates from t-SNE, observations were plotted as markers on a scatter plot (\textit{plt.scatter} from \textit{matplotlib}).
Markers were color-coded based on ground truth labels. Class and feature centroids were plotted, color-coded, and annotated by the same methods.

\begin{tcolorbox}[sharp corners, boxrule=0.5pt]
    \footnotesize

        \begin{minipage}{0.8\textwidth}
        \footnotesize
        Given a data array $\textbf{D}$, with a set of features \textbf{F} and class labels \textbf{L}.\\
        \textbf{Class Centroid Calculation:}\\
        For class $c$ in \textbf{L}, \\
        where $N_c$ is the number of observations in class $c$:
        \begin{equation}
        \text{Class Centroid (C}_x, C_y) = \left(\frac{\sum x_i}{N_c}, \frac{\sum y_i}{N_c}\right)
        \label{equation_class}
        \end{equation}
        where $(x_i, y_i)$ are the coordinates of observations in class $c$.
        
        \textbf{Feature Centroid Calculation:}\\
        For feature $f$ in \textbf{F}, \\
        where $N_f$ is the number of observations with feature $f$:
        \begin{equation}
        \text{Feature Centroid} (F_x, F_y) = \left(\frac{\sum x_i}{N_f}, \frac{\sum y_i}{N_f}\right)
        \label{equation_feature}
        \end{equation}
        where $(x_i, y_i)$ are coordinates of observations with feature $f$.
        \end{minipage}

    \end{tcolorbox}

\begin{figure}
    \centering
    \includegraphics[width=0.47\textwidth]{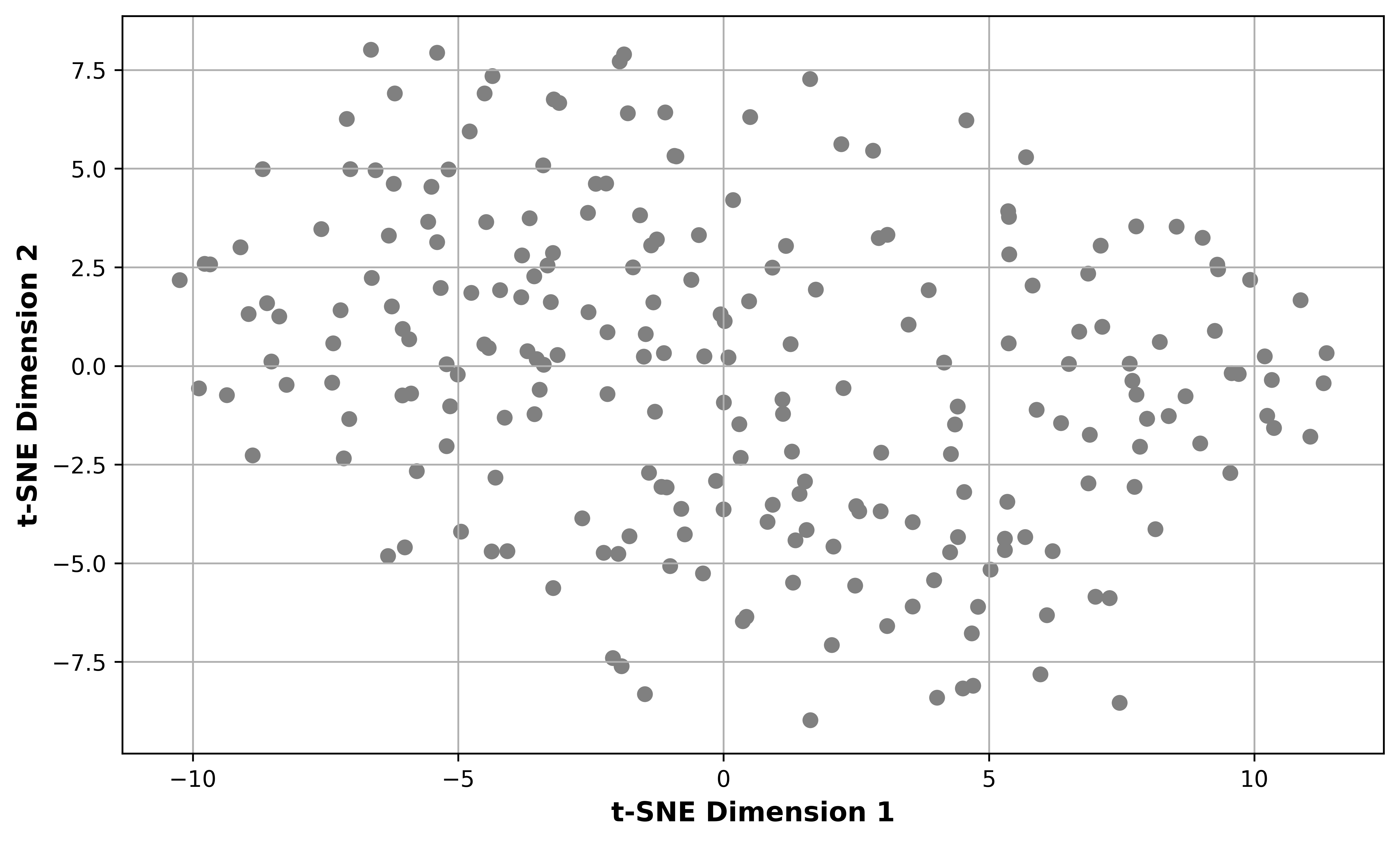}
    \captionsetup{font=footnotesize}
    \caption{
        \footnotesize{Scatter plot of 235 neurogenetic disease variants in two-dimensional space. The \textit{x} and \textit{y} coordinates were calculated by t-SNE based on the 31 phenotype features.}}
    \label{fig:No_colors}
\end{figure}

\begin{figure}
\centering
\includegraphics[width =0.47\textwidth]{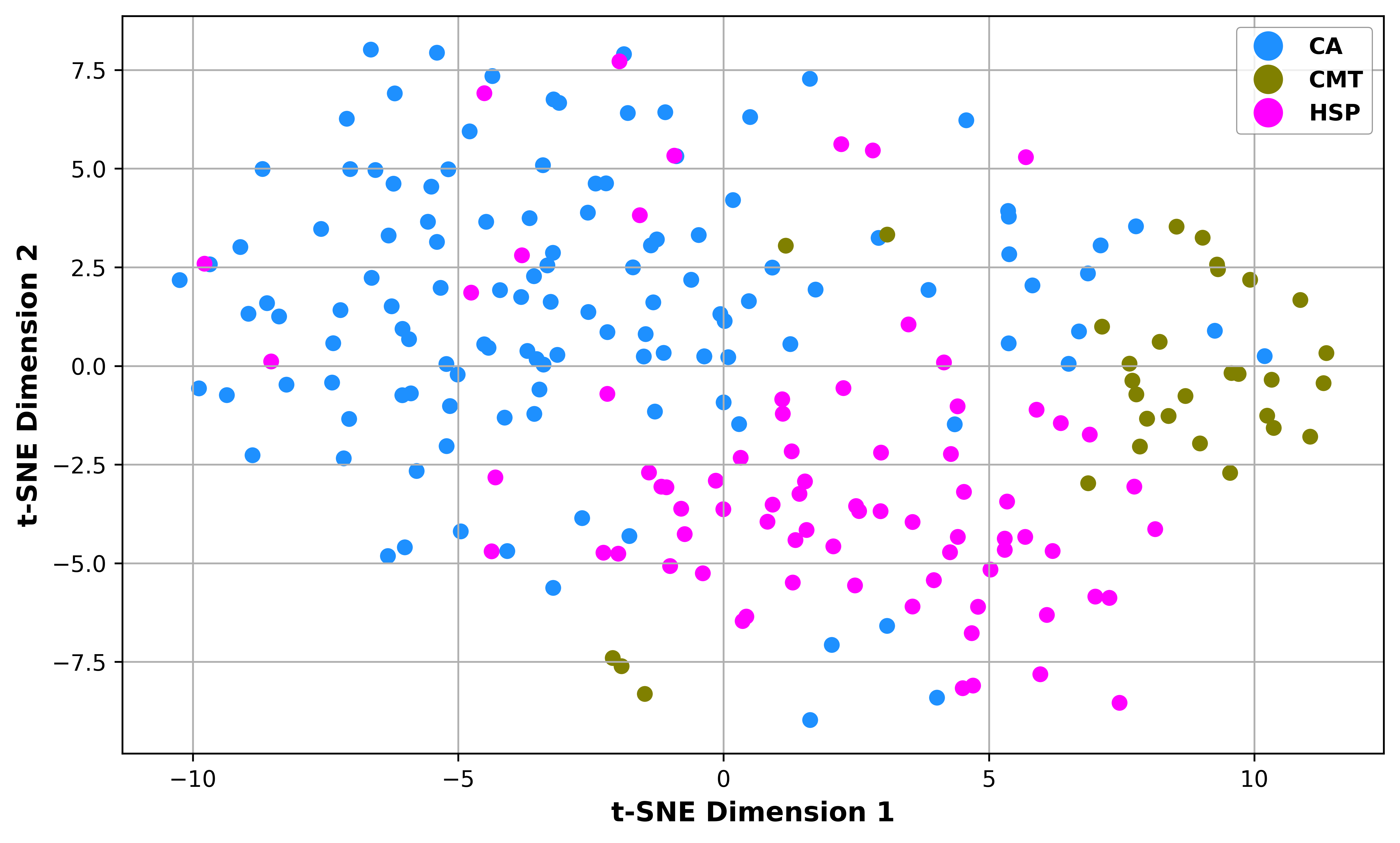}
\captionsetup{font=footnotesize}
\caption{Same scatter plot as Fig. \ref{fig:No_colors} with markers colored by their ground truth labels. Note that markers form three groupings based on their class membership. There are 33 variants of Charcot-Marie-Tooth disease, 77 variants of hereditary spastic paraparesis, and 125 variants of cerebellar ataxia shown.  Each variant has a distinct phenotype.}
\label{fig:Colors}
\end{figure}

\begin{figure}
\centering
\includegraphics[width=0.47\textwidth]{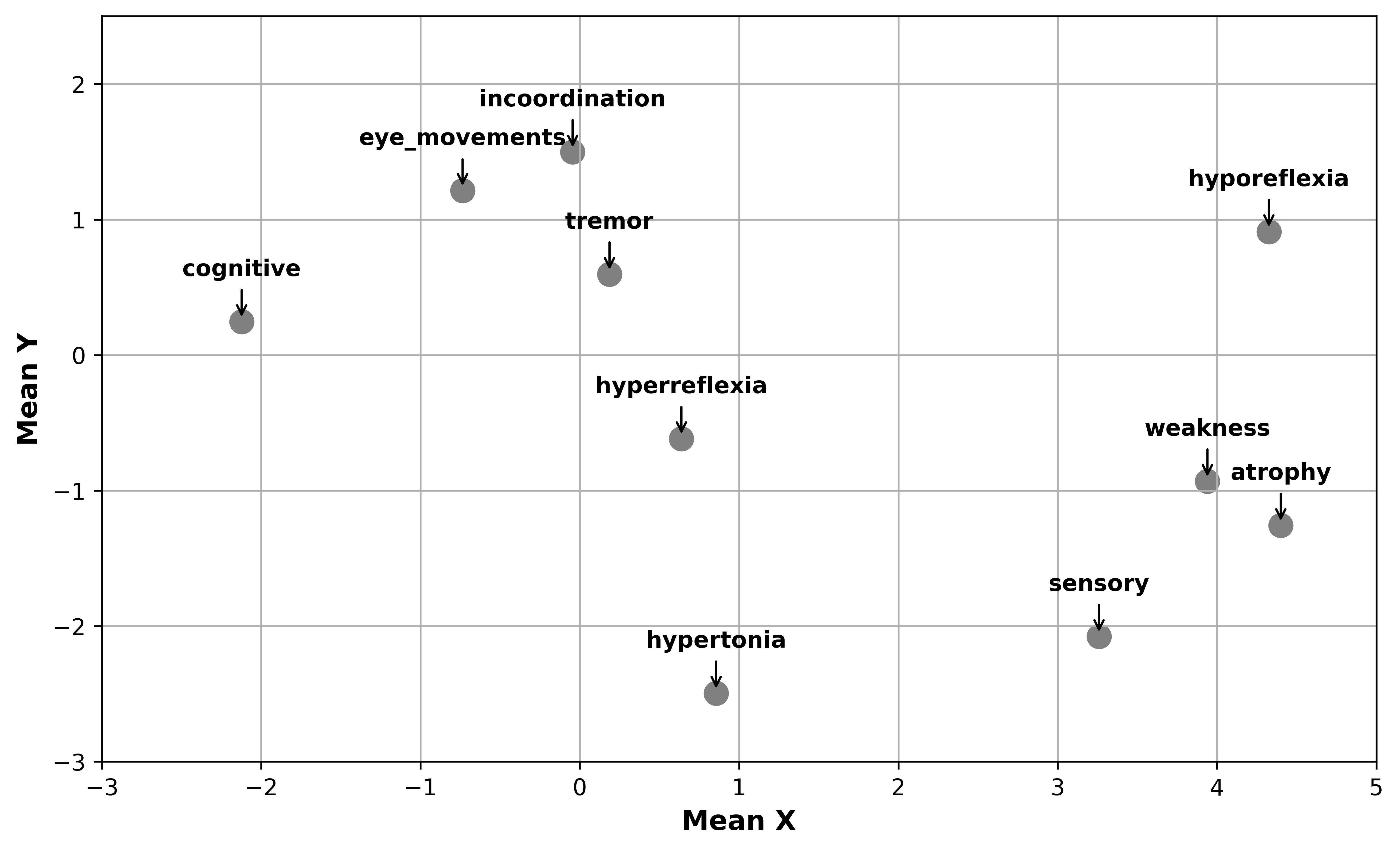}
\captionsetup{font=footnotesize}
\caption{Feature centroids for ten of the most important features. The x- and y-coordinates of these features have been calculated based on the x- and y-coordinates of the disease variants that exhibit these features (phenotypes). See Methods for details.}
\label{fig:Features}
\end{figure}

\begin{figure}
    \centering
    \includegraphics[width=0.47\textwidth]{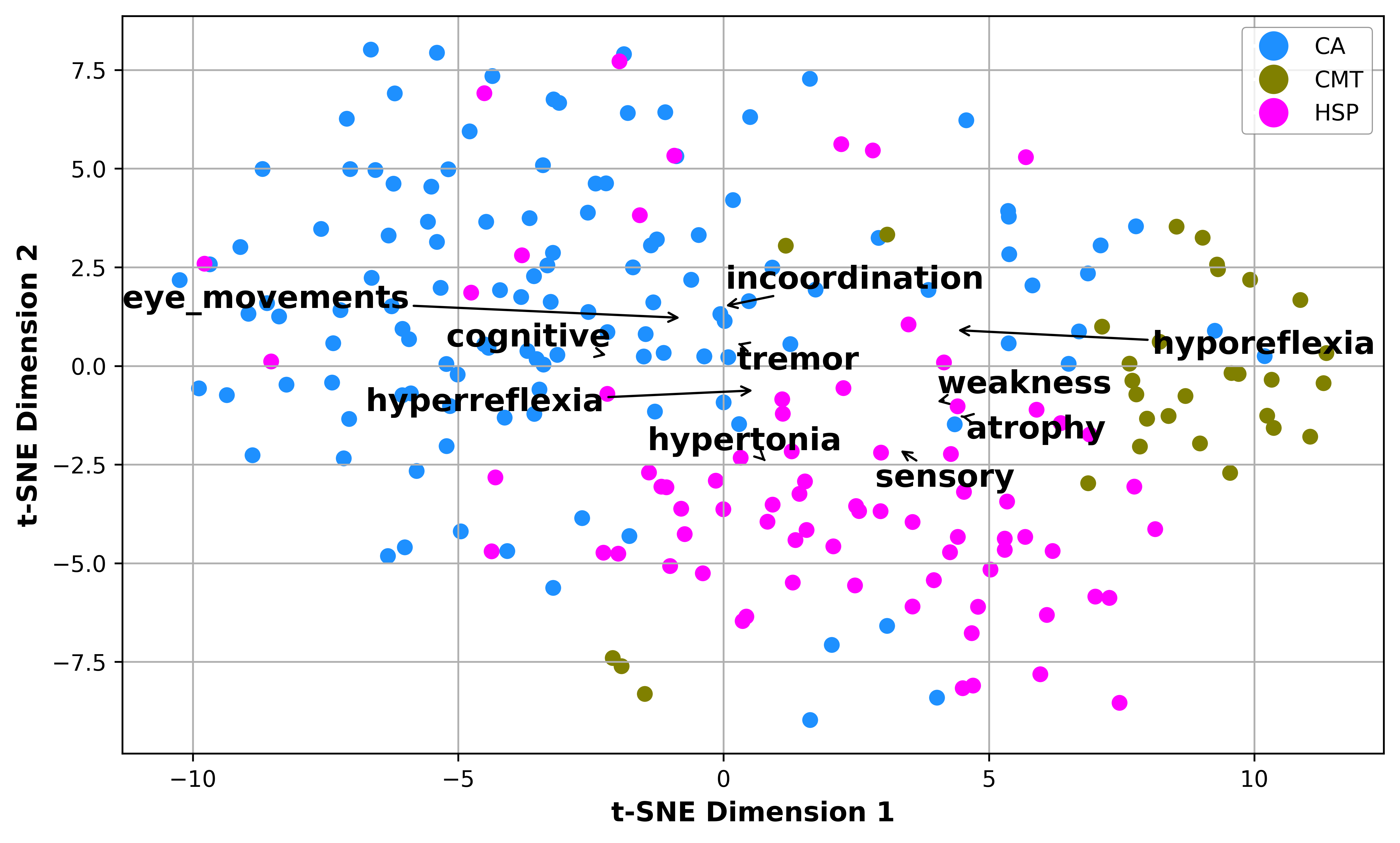}
    \captionsetup{font=footnotesize}
    \caption{
        \footnotesize{
            Same data as Fig. \ref{fig:Colors}. Feature centroids have been added to the scatter plot.  Although the scatter plot is cluttered, note the proximity of Charcot-Marie-Tooth disease to feature centroids for weakness, atrophy, and hyporeflexia; the proximity of cerebellar ataxia markers to the feature centroids for eye movements, incoordination, tremor, and cognitive; and the proximity of hereditary spastic paraparesis markers to hypertonia, hyperreflexia, and sensory feature centroids.}}
    \label{fig:Markers with feature centroids}
\end{figure}

\begin{figure}
    \centering
    \includegraphics[width=0.47\textwidth]{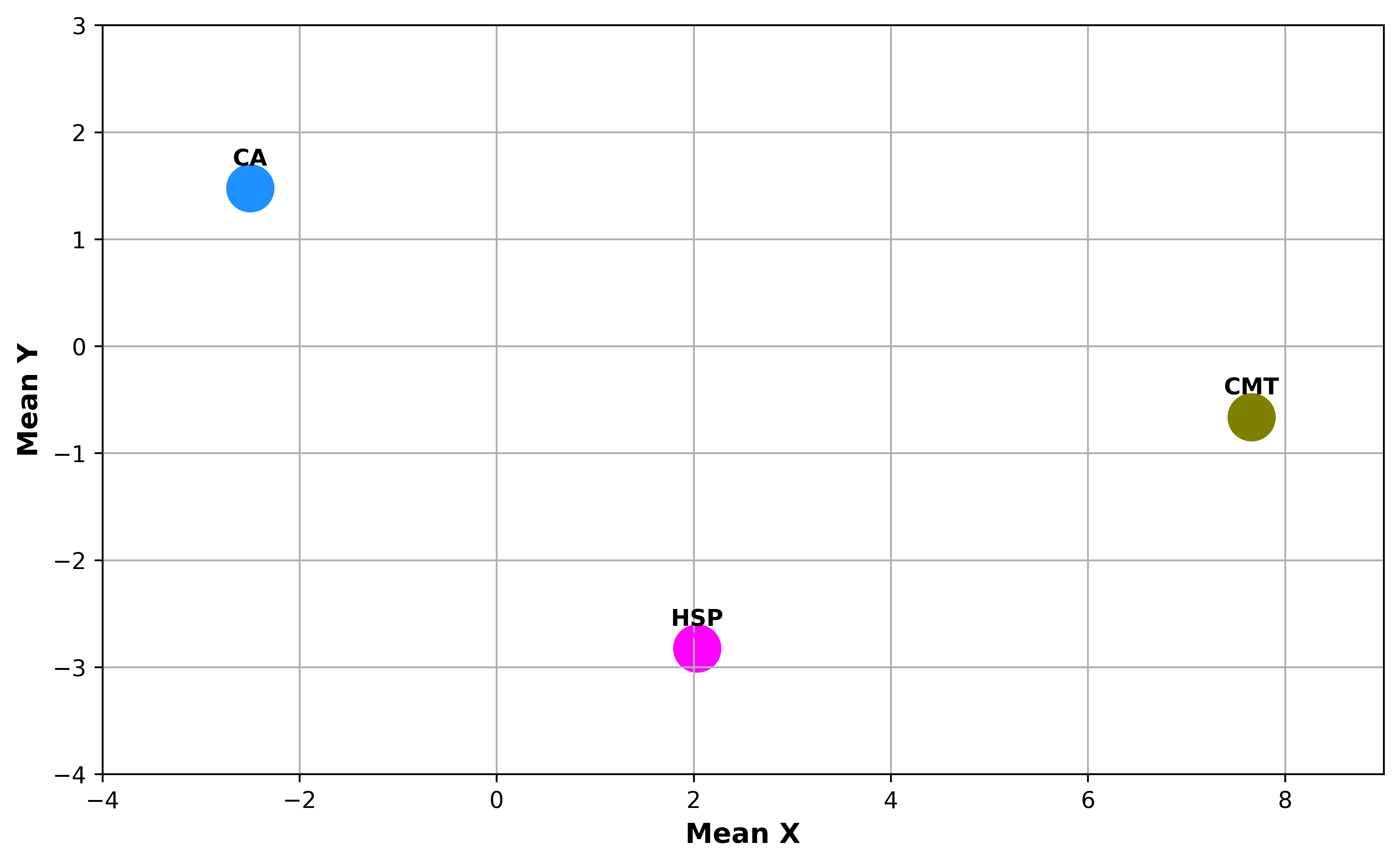}
    \captionsetup{font=footnotesize}
    \caption{
        \footnotesize{Class centroids for the three neurogenetic disease classes, compared
        to Fig. \ref{fig:Colors}.
        }
    }    
    \label{fig:Classes}
\end{figure}

\begin{figure}
\centering
\includegraphics[width =0.47\textwidth]{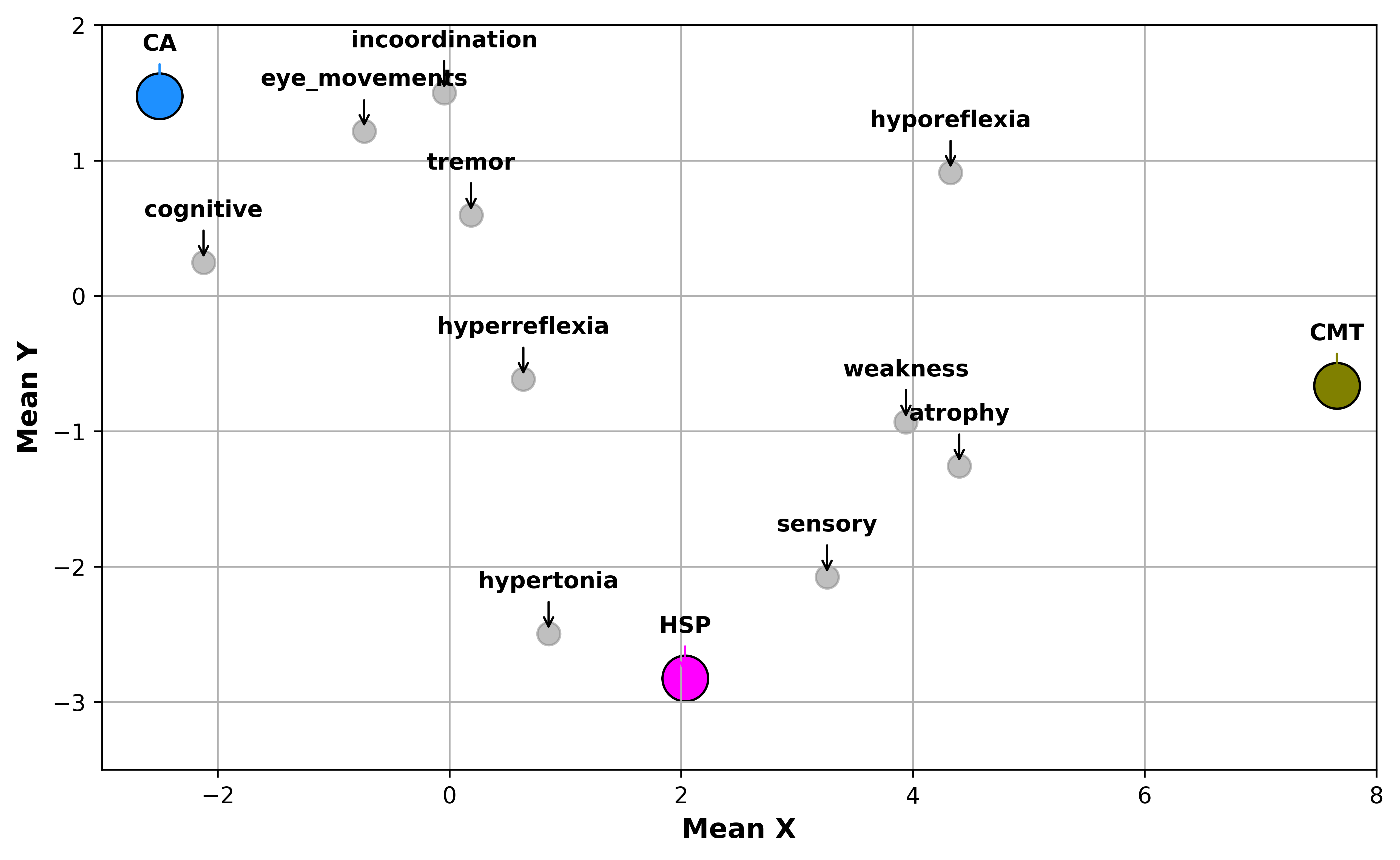}
\captionsetup{font=footnotesize}
\caption{\footnotesize{Composite scatter plot that combines the feature centroids from Fig. \ref{fig:Features} with the class centroids from Fig. \ref{fig:Classes}.}}
\label{fig:Composite centroids}
\end{figure}
\section{Results}
\label{s:results}
We downloaded the phenotypes of 235 neurogenetic disease variants from Orphadata. Orphadata encodes phenotypes with terms from the Human Phenotype Ontology \cite{robinson2012deep}. Orphadata uses $970$ unique terms from the HPO to describe the phenotypes of these disease variants. We used reduction by subsumption to collapse these 970 terms into 31 phenotype categories (Figs. \ref{fig:Subsumption} and \ref{fig:bar_chart_neurogenetic}). We used t-SNE to reduce the data to two dimensions. This approach provided \textit{x} and \textit{y} coordinates for each observation, enabling us to calculate the \textit{x,y} coordinates of each observation (disease variant) as well as the mean \textit{x,y} coordinates of the class and feature centroids.

The 235 observations can be represented as markers on a scatter plot (Fig. \ref{fig:No_colors}). Since the ground truth label is available for each observation, each marker can be color-coded by class name (Fig. \ref{fig:Colors}).  
We calculated the feature centroids for each of the 31 phenotype categories (Equation \ref{equation_feature}) and the class centroids for the three disease classes (Equation \ref{equation_class}). Rather than plot all 31 feature centroids on a scatter plot, we used  Shapley Additive exPlanations (SHAP) to find the most informative features (sometimes influential or important) (Fig. \ref{fig:SHAP}). 
 The centroids of the 10 most informative phenotype features are shown in Fig. \ref{fig:Features}.  These feature centroids were overlaid onto a scatter plot with all 235 observations (Fig. \ref{fig:Markers with feature centroids}). 
 We plotted class centroids (Fig. \ref{fig:Classes}) and then combined the feature centroids and the class centroids into a single plot (Fig. \ref{fig:Composite centroids}).

\section{Discussion}
\label{s:discussion}
Dimension reduction is a powerful tool for gaining insight into high-dimension data. As Rudin et al. \cite{rudin2022interpretable} commented, \textit{even in data science, a picture is worth a thousand words. Dimension reduction (DR) techniques take, as input, high-dimensional data and project it down to a lower-dimensional space (usually 2D or 3D) so that a human can better comprehend it.} However, interpreting low-dimension representations of high-dimension spaces is difficult.  The x- and y-axis of two-dimensional spaces derived from dimension reduction are in arbitrary units and have no simple interpretation.  Rudin et al.\cite{rudin2022interpretable} consider the interpretation of spaces created by dimension reduction one of the grand challenges of explainable AI and ask \textit{Can the DR transformation from high- to low-dimensions be made more interpretable or explainable?} We suggest that the overlay of class and feature centroids 
onto dimension-reduced scatter plots more interpretable. 
When datasets are reduced to two dimensions, \textit{x} and \textit{y} coordinates allow each observation to be represented as a marker on a scatter plot (Figs. \ref{fig:No_colors} and \ref{fig:Colors}).  These x and y coordinates also allow a calculation of feature and class centroids which can be overlaid onto scatter plots (Figs. \ref{fig:Classes} and \ref{fig:Features}).

The overlay of feature and class centroids onto dimension-reduced scatter plots offers new avenues for interpretation (Fig. \ref{fig:Composite centroids}). One approach is to examine scatter plots on a quadrant-by-quadrant basis and look for classes and features that share the same quadrant.  For example, in Fig. \ref{fig:Composite centroids}, the upper left quadrant is inhabited by the class CA (cerebellar ataxia) and the features incoordination, cognitive, tremor, hyperreflexia, and incoordination. 
Importantly, tremor, eye movement abnormalities, and incoordination are key presenting signs of cerebellar ataxia.  Another approach is to use the proximity between class and feature centroids to interpret the dimension-reduced scatter plot.  For example, the centroid for hereditary spastic paraparesis (HSP) is closest to the features of hypertonia, sensory loss, atrophy, weakness, and hyperreflexia (Fig. \ref{fig:Composite centroids}), and signs and symptoms of this disorder.


Although our novel approach, which relies on calculating class and feature centroids, appears promising, it does have limitations. Since most dimension reduction methods are nonlinear, there is no straightforward way to interpret distances between centroids on dimension-reduced plots (Figs. \ref{fig:Features} and \ref{fig:Classes}). The relative distances between centroids must be interpreted with caution.
Second, more work needs to be done on scalability.
Our studies were carried out on a small $235 \times 31$ data array; additional work on larger datasets is needed.
Third, more work needs to be done on generalizability.
Our dataset was based on neurological phenotypes that were binary-coded. Further work can assess whether the same methods can be applied to continuous or ordinal data from other domains of interest.

Nevertheless, this approach to improving the interpretability dimension-reduced scatter plots has advantages. First, adding feature centroids provides a link to the underlying feature space, aiding in understanding the significance of the \textit{x} and \textit{y} axes. Second, replacing observation markers with class centroids improves readability (Fig. \ref{fig:Classes}). Third, the calculation of feature and class centroids is straightforward and the resulting centroids are easily plotted using \textit{matplotlib}.
This method could be useful in biology, medicine, and bioinformatics, where dimension-reduced scatter plots are widely used.

\section*{Data and code availability}
\label{s:data-code}
Data for the neurogenetic diseases and Python code are available on the project GitHub site at \textbf{{https://github.com/ACIL-Group/CENTROIDS} }and archived at Zenodo \cite{CENTROIDS-Zenodo}.
\bibliographystyle{IEEEtran}
\bibliography{references}
\end{document}